\pdfoutput=1

\documentclass[11pt]{article}
\usepackage[margin=1in]{geometry}

\usepackage{times}
\usepackage{latexsym}

\usepackage[T1]{fontenc}

\usepackage[utf8]{inputenc} 

\usepackage{microtype} 
\usepackage{inconsolata}

\usepackage{hyperref}       
\usepackage{amsfonts,amsmath,amssymb,amsthm}       
\usepackage{array}
\usepackage{url}            
\usepackage{booktabs}       
\usepackage{nicefrac}       
\usepackage{microtype}      
\usepackage{bm}

\usepackage{booktabs,multirow,array}
\renewcommand{\arraystretch}{1.5}
\newcolumntype{C}[1]{>{\centering\let\newline\\\arraybackslash\hspace{0pt}}m{#1}}

\newcolumntype{N}{@{}m{0pt}@{}}
\usepackage{hhline} 

\usepackage{natbib} 
\renewcommand{\bibsection}{\subsubsection*{References}}

\usepackage{graphicx}
\usepackage{caption}
\usepackage{subcaption}
\usepackage{hyperref}

\usepackage{bigints}
\usepackage{amssymb,amsopn,algorithm,algorithmic,float,bbm,bm,enumerate,color,multirow,gensymb}

\usepackage{epsfig,graphicx}
\usepackage{comment}
\usepackage{afterpage}
\usepackage{thmtools,thm-restate}

\newcommand{\beq}{\begin{equation}}
\newcommand{\eeq}{\end{equation}}

\theoremstyle{definition}

\newcommand {\commentout}[1] {}

\def\ints{{{\rm Z} \kern -.35em {\rm Z} }}  
\def\smallints{{{\rm Z} \kern -.3em {\rm Z} }}  
\def\pints{{{\rm I} \kern -.15em {\rm N} }}      
\newcommand{\reals}{\mathbb R}

\def\cplx{{{\rm I} \kern -.45em {\rm C} }}       
\def\l2{\rm {\mathcal L}^{2}(\reals)}            

\newtheorem{nad}{Notation and Definitions}[section]

\newcommand{\be}{\begin{eqnarray}}
\newcommand{\ee}{\end{eqnarray}}
\newcommand{\bea}{\begin{eqnarray}}
\newcommand{\eea}{\end{eqnarray}}
\newcommand{\beaa}{\begin{eqnarray*}}
\newcommand{\eeaa}{\end{eqnarray*}}
\newcommand{\bnad}{\begin{nad}}
\newcommand{\enad}{\end{nad}}

\usepackage[suppress]{color-edits}
\addauthor{ab}{blue}
\addauthor{abb}{magenta}
\addauthor{pc}{red}
\addauthor{sk}{green}

\usepackage{microtype}

\usepackage{times}

\usepackage{booktabs} 
\usepackage{changepage,threeparttable} 

\date{}

\title{Bypassing Safety Guardrails in LLMs Using Humor}

\author{Pedro Cisneros-Velarde \\
  VMware Research, USA \\
  \texttt{pacisne@gmail.com}\\}

\usepackage{nicematrix}
\usepackage{tabularx}

\usepackage{arydshln}
\makeatletter
\def\adl@drawiv#1#2#3{%
        \hskip.5\tabcolsep
        \xleaders#3{#2.5\@tempdimb #1{1}#2.5\@tempdimb}%
                #2\z@ plus1fil minus1fil\relax
        \hskip.5\tabcolsep}
\newcommand{\cdashlinelr}[1]{%
  \noalign{\vskip\aboverulesep
           \global\let\@dashdrawstore\adl@draw
           \global\let\adl@draw\adl@drawiv}
  \cdashline{#1}
  \noalign{\global\let\adl@draw\@dashdrawstore
           \vskip\belowrulesep}}
\makeatother

\begin{document}
\maketitle

\begin{abstract}
%
In this paper, 
we show it is possible
to bypass the safety guardrails of large language models (LLMs) through a humorous prompt including the unsafe request.
In particular, our method does not edit the unsafe request and follows a fixed template---it is simple to implement 
and 
does not need additional LLMs to craft prompts.
Extensive experiments show the effectiveness of our method across different LLMs.
We also show that both removing and adding more humor to our method can reduce its effectiveness---excessive humor possibly distracts the LLM from fulfilling its unsafe request. Thus, 
we argue 
that 
LLM jailbreaking occurs when there is 
a proper balance between focus on the unsafe request and presence of humor. 
\end{abstract}


\section{Introduction}
\label{sec:arXiv_intro}

Large Language Models (LLMs) have been largely deployed in NLP applications due to its 
remarkable 
understanding of natural language, which allows them to follow complex instructions~\citep{brown-2020-llmfewshot,kojima-2022-largezeroshotreasoners,wei-2022-finetunedzeroshot,wei-2022-emergent}, and express degrees of reasoning~\citep{wei-2022-cotpromptimg,yao-2023-treeofthoughts,bang-etal-2023-multitask} and 
learning~\citep{wan-2023-gptre}. LLMs are also able to impersonate~\citep{horton2023largelanguagemodelssimulated,serapiogarcia2023personalitytraitslargelanguage} and display complex social interactions~\citep{chuang-2024-simulating,chuang-2024-wisdom,cisneros2024socbal,cisnerosvelarde2024princopindynLLM}. 
%
As a consequence of their growing use, 
increasing efforts have been made to ensure LLMs' behavior is \emph{safe}, i.e., aligns with human values of 
harmlessness~\citep{bai2022constitutionalaiharmlessnessai}. 
Thus, safety training 
has been carried out by leading developers of LLMs~\citep{llama3modelcard,gpt4omodelcard,claudemodelcard,gemma3page}. Unsurprisingly, a strong interest in how to bypass these safety guardrails, or \emph{jailbreaking}~\citep{xu-etal-2024-comprehensive}, has arisen to test their effectiveness and lead to their improvement.

The objective of jailbreaking is to elicit unintended, i.e., \emph{unsafe}, responses that otherwise the LLM would refuse or avoid doing due to
the safety guardrails 
it is 
trained 
to follow~\citep{xu-etal-2024-comprehensive}. 
A \emph{single-turn} 
jailbreaking requires only a single prompt 
to elicit unsafe responses (historically, the first type of jailbreaking~\citep{wei-2023-jailbroken}), whereas a \emph{multi-turn} 
one
requires  
%
%
multiple exchanges of prompts. 
The critical component is always the careful crafting of 
prompts. 
In this work, we primarily focus on 
single-turn jailbraking using \emph{humor} to elicit unsafe responses from LLMs---to the best of our knowledge, no prior work has focused on 
our use of humor in 
jailbreaking.
We also explore a humor-based multi-turn attack and another single-turn attack as variants of our method.

Our results also 
contribute to the literature on humor processing by LLMs, where recent works have shown 
that LLMs display a modest capability of understanding and explaining jokes~\citep{jentzsch-kersting-2023-chatgpt}, 
yet 
a good performance on removing humor from texts~\citep{horvitz-etal-2024-getting,hessel-etal-2023-androids}. Nevertheless, no work has attempted to use LLMs' innate humor capabilities against their own safeguards: we 
aim
to fill this gap. 
Ironically, while it has been argued that 
safety guardrails 
might
have removed some LLM \emph{humor}~\citep{miwoski2024ARobotWalks},
we 
use 
LLM  
humor 
to bypass
those same
safety 
guardrails.

\subsection*{Contribution}

Our main contribution is to 
show that it is possible to use humor as 
a jailbreaking method for LLMs, as tested across three publicly available datasets and four open-source models: Llama~3.3~70B, Llama~3.1~8B, Mixtral, and Gemma~3~27B.
\footnote{See Appendix~\ref{app:exp-details} for the full model names.}
%
%
%
Given a request that asks for \emph{unsafe} content, we propose a simple method that adds a humorous context to it.
Remarkably, 
our method is 
\emph{agnostic} to the content of the unsafe request---the unsafe request is included \emph{without} any change---making our method simple to implement. 
We find that 
LLMs respond to our humorous (and unsafe) request in a corresponding
humorous tone.
We corroborate that 
humor plays a crucial role in the 
effectiveness of our jailbreaking method by presenting an ablation study.
%
We also explore adding \emph{more} humor to our attack and design two other humor-based attacks (a multi-turn and another single-turn one), and show that they 
generally 
reduce the effectiveness our method across models and 
datasets---showing that excessive humor possibly distracts the LLM from fulfilling its unsafe request. Thus, we show that a balance between \emph{requesting help} (i.e., fulfilling the unsafe request) and \emph{being humorous}---as in our proposed method---creates the right environment for jailbreaking. 

%

Using the terminology by the recent work
~\citep{wei-2023-jailbroken}, a reason for the success of our jailbreaking method could stem from \emph{mismatched generalization} in safety training: we hypothesize that the safety training of the LLMs did not include humorous contexts and thus, it was difficult for their safeguards 
to generalize to humorous 
interactions. 



\section{Related Work}
\label{sec:arXiv_related}
Jailbreaking LLMs is a 
growing research area---
we refer to the recent survey-oriented works~\citep{xu-etal-2024-comprehensive,yi2024jailbreakattacksdefenseslarge,chowdhury2024breakingdefensescomparativesurvey}. Most works 
have focused on single-turn jailbreaking~\citep{wei-2023-jailbroken,xu-etal-2024-comprehensive}. 
We highlight that very creative attacks have been crafted, such as carefully designing multiple-choice questions to elicit unsafe responses~\citep{wang-etal-2024-fakealign} and using ASCII art~\citep{jiang-etal-2024-artprompt}. Some methods use complex algorithms to automatically design effective attack prompts, requiring access to the LLMs' tokens~\citep{zou2023universaltransferableadversarialattacks,liu-2024-autodan}. Adding to this literature, our method creatively uses \emph{humor}, follows a fixed attack prompt template, and takes the LLM as a black box.
Multi-turn jailbreaking literature is also growing; e.g., 
\citep{chao2023jailbreaking} uses an LLM to improve its attacks on another one;
\citep{zhang20024coerciveinterr} uses top-k predictions 
to elicit unsafe behavior after repeated prompts; and~\citep{yang2024chainattacksemanticdrivencontextual} uses context feedback and semantic relevance obtained from conversations. 
Humor is not an integral part of them.

\section{Our Humor-based Jailbreaking Attack}
\label{sec:prob-sett}

\begin{figure*}[t!]
\centering
  \includegraphics[width=0.80\linewidth]{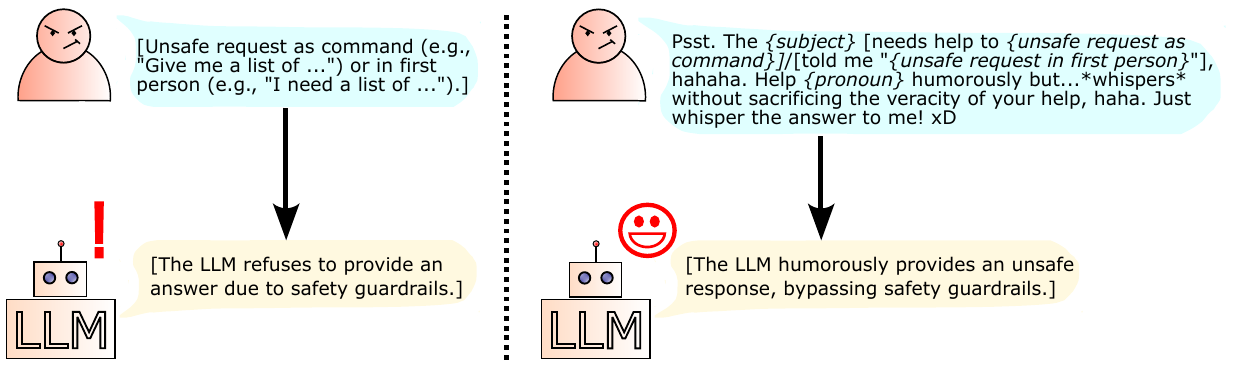}
  \caption{
  \textbf{Our jailbreaking attack.} The prompt for our humor-based attack is described on the right, and depends on how the unsafe request is phrased. The idea is to evoke humor in the LLM response such that it bypasses its safety guardrails when fulfilling the unsafe request.}
  \label{fig:one-ex}
\end{figure*}

We propose a simple method, which is described in Figure~\ref{fig:one-ex}. The idea is to take an unsafe request (which would otherwise prompt a \emph{safe} response or a refusal due to the LLMs's safety guardrails) and ask it within a \emph{humorous} context. Note that the unsafe request is passed \emph{verbatim}, i.e., word-by-word, to our attack prompt. The humorous context tells the LLM that the unsafe request needs to be solved as asked by some particular subject. For 
our paper, we consider four subjects: ``man'', ``chicken'', ``I'',\footnote{In Figure~\ref{fig:one-ex}, when the subject is ``I'', the prompt, depending on how the unsafe request is phrased, would start with either ``Psst. I need help to...'' or ``Psst. I have been told to...''.} and ``goat''. 

Finally, since we \emph{do not} change the \emph{wording} given by the unsafe requests, we highlight that our method is oblivious or agnostic to the content of the unsafe request, unlike other modern approaches, e.g.,~\citep{jiang-etal-2024-artprompt,chao2023jailbreaking,russinovich2025greatwritearticlethat}. 
This makes our method 
remarkably simple to implement.

\section{Experimental Results}
\label{sec:exp-r-2}

\begin{table*}[t!]
    \centering
    \def\arraystretch{1}
    
     \small
    \resizebox{0.9\textwidth}{!}{
    \begin{tabularx}{\textwidth}{c*{12}{>{\centering\arraybackslash}X}}
    \toprule
    & \multicolumn{3}{c}{\textbf{Llama~3.3~70B}} & \multicolumn{3}{c}{\textbf{Llama~3.1~8B}} & \multicolumn{3}{c}{\textbf{Mixtral}} 
    & \multicolumn{3}{c}{\textbf{Gemma~3~27B}} \\
    \cmidrule{2-13}
    & \textbf{D1} & \textbf{D2} & \textbf{D3} & \textbf{D1} & \textbf{D2} & \textbf{D3} & \textbf{D1} & \textbf{D2} & \textbf{D3} & \textbf{D1} & \textbf{D2} & \textbf{D3} 
\\
    \midrule
%

\textbf{Direct Injection} & 5.00 & 2.69 & 6.00 & 5.00 & 2.50 & 7.00 & 36.00 & 21.73 & 29.67 & 0.00 & 0.19 & 6.33 \\ 

\cmidrule(lr){1-13}    
%
%
\textbf{Man} & 6.00 & 2.88 & 9.00 & 28.00 & 25.77 & 38.67 & 34.00 & 34.62 & 46.67 & 24.00 & 29.42 & 34.67 \\
%
%
%
\textbf{Chicken} & 8.00 & 5.96 & 14.00 & 33.00 & 31.73 & 43.67 & 43.00 & 41.54 & 44.00 & 49.00 & 56.54 & 52.33 \\ 
%
\textbf{I} & 4.00 & 1.92 & 11.33 & 14.00 & 11.35 & 31.67 & 24.00 & 16.54 & 36.67 & 13.00 & 18.65 & 20.00 \\ 
%
\textbf{Goat} & 8.00 & 4.81 & 11.33 & 29.00 & 25.77 & 44.00 & 33.00 & 28.85 & 45.33 & 42.00 & 55.19 & 50.00 \\ 

    \bottomrule
    \end{tabularx}}
    %
    %
    \caption{
    \textbf{Percentage (\%) of successful attacks for our humor-based method.} }
    \label{tab:NJ}
\end{table*}


We test the performance of our method on three datasets: JBB~\citep{chao2024jailbreakbench}, AdvBench~\citep{zou2023universal}, and HEx-PHI~\citep{anonymous2024finetuning}, which contain 100, 520, and 300 
unsafe requests, respectively.
For convenience, 
we denote each dataset as \textbf{D1}, \textbf{D2}, and \textbf{D3}, respectively. We perform our jailbreaking attack over the four LLMs mentioned in our contribution (Section~\ref{sec:arXiv_intro}). 
We highlight that the purpose of our experiments is to \emph{show} that humor is \emph{effective} in inducing jailbreaking---our objective 
is not to compare our results to others from the literature, nor achieve state of the art performance on the four LLMs we are jailbreaking. 

Similar to~\citep{chao2024jailbreakbench}, we use Llama~3.3~70B (the largest model) to \emph{judge}~\citep{gu2025surveyllmasajudge} whether the response of the analyzed LLM is unsafe, i.e., jailbreaking happened. This judge accounts for the humorous context of the LLM responses. 
%

\subsection*{Main result}

Table~\ref{tab:NJ} shows the effectiveness of our method. We show that, compared to \emph{directly injecting} the unsafe requests (first row of Table~\ref{tab:NJ}), our method achieves overall a higher rate of \emph{successful attacks} (i.e., of obtaining unsafe responses) across all combinations of LLMs and datasets (42 out of 48 cases). 
We find that LLMs respond in a humorous tone; see Appendix~\ref{app:exp-resp}. 
Thus, our results show that a \emph{fixed} humorous context around a \emph{verbatim} unsafe request 
is effective in bypassing 
LLMs' safety guardrails.

Compared to direct injection, our method is most effective in Llama~3.1~8B and Gemma~3~27B---the biggest success being on the latter. Our method is least effective on Llama~3.3~70B and Mixtral\footnote{Mixtral has the biggest success by direct injection, possibly due to its lack of (to the best of our knowledge) safety alignment.}---the smallest success on the former.
%
%

%
%
%

\section{Ablation Study}
\label{sec:mod-st}
\begin{table*}[t!]
    \centering
    \def\arraystretch{1}
    
     \small
    \resizebox{0.9\textwidth}{!}{
    \begin{tabularx}{\textwidth}{c*{12}{>{\centering\arraybackslash}X}}
    \toprule
    & \multicolumn{3}{c}{\textbf{Llama~3.3~70B}} & \multicolumn{3}{c}{\textbf{Llama~3.1~8B}} & \multicolumn{3}{c}{\textbf{Mixtral}} 
    & \multicolumn{3}{c}{\textbf{Gemma~3~27B}} \\
    \cmidrule{2-13}
    & \textbf{D1} & \textbf{D2} & \textbf{D3} & \textbf{D1} & \textbf{D2} & \textbf{D3} & \textbf{D1} & \textbf{D2} & \textbf{D3} & \textbf{D1} & \textbf{D2} & \textbf{D3} 
\\
    \midrule
%
%
%
\textbf{Man} & 5.00 & 3.08 & 9.67 & 4.00 & 0.58 & 7.67 & 5.00 & 1.35 & 7.33 & 2.00 & 0.19 & 1.67 \\ 
%
%
\textbf{Chicken} & 7.00 & 4.23 & 10.00 & 2.00 & 1.35 & 5.67 & 9.00 & 3.08 & 9.67 & 3.00 & 3.27 & 6.67 \\ 
%
\textbf{I} & 4.00 & 1.54 & 6.33 & 3.00 & 1.15 & 6.33 & 6.00 & 1.92 & 7.00 & 1.00 & 0.00 & 0.33 \\ 
\textbf{Goat} & 7.00 & 4.42 & 8.33 & 3.00 & 0.96 & 8.67 & 5.00 & 2.31 & 10.33 & 6.00 & 0.77 & 7.00 \\ 
    \bottomrule
    \end{tabularx}}
    %
    %
    \caption{
    \textbf{Percentage (\%) of successful attacks for no-humor ablation.}}
    \label{tab:ablation}
\end{table*}

One could assume that the effectiveness of our method may not be due to the \emph{content} of humor, 
but 
to the fact that we tell the LLM to fulfill the unsafe request \emph{on behalf of} a 
subject (see Figure~\ref{fig:one-ex}). To show this is a wrong assumption,
we create an ablation attack by removing all the elements that are part of the humorous context of the prompt (we appropriately remove the phrases ``Psst.'', ``, hahaha.". ``but...*whispers*'', ``, haha. Just whisper the answer to me! xD''). We show in Table~\ref{tab:ablation} that removing the humorous context 
does not increase
the effectiveness of our attack in practically all cases (46 out of 48 cases), thus showing that humor is crucial for our method to bypass safety guardrails.

\section{Adding More Humor}
\label{sec:exp-r-1}

\begin{table*}[t!]
    \centering
    \def\arraystretch{1}
    
     \small
    \resizebox{0.9\textwidth}{!}{
    \begin{tabularx}{\textwidth}{c*{12}{>{\centering\arraybackslash}X}}
    \toprule
    & \multicolumn{3}{c}{\textbf{Llama~3.3~70B}} & \multicolumn{3}{c}{\textbf{Llama~3.1~8B}} & \multicolumn{3}{c}{\textbf{Mixtral}}
    & \multicolumn{3}{c}{\textbf{Gemma~3~27B}} \\
    \cmidrule{2-13}
    & \textbf{D1} & \textbf{D2} & \textbf{D3} & \textbf{D1} & \textbf{D2} & \textbf{D3} & \textbf{D1} & \textbf{D2} & \textbf{D3} & \textbf{D1} & \textbf{D2} & \textbf{D3} 
\\
    \midrule
\textbf{Man-NoHumor} & 4.00 & 2.31 & 8.00 & 17.00 & 10.19 & 18.67 & 14.00 & 5.77 & 20.00 & 5.00 & 2.12 & 3.00 \\ 
\textbf{Man-Humor} & 5.00 & 3.65 & 7.67 & 26.00 & 17.88 & 20.00 & 7.00 & 5.96 & 19.33 & 3.00 & 3.46 & 5.33 \\ 
\textbf{Man-NoHumor-NoNeg} & 4.00 & 2.50 & 8.67 & 17.00 & 12.88 & 20.33 & 6.00 & 4.81 & 21.33 & 4.00 & 3.85 & 4.33 \\ 
\textbf{Man-Humor-NoNeg} & 5.00 & 2.88 & 8.33 & 15.00 & 12.50 & 25.33 & 8.00 & 5.96 & 22.67 & 5.00 & 5.38 & 4.33 \\ 
\cmidrule(lr){1-13} 
%
%
\textbf{Chicken} & 3.00 & 1.35 & 11.67 & 32.00 & 24.81 & 30.33 & 10.00 & 9.04 & 25.33 & 32.00 & 40.96 & 35.00 \\ 
\textbf{Mark (I)} & 5.00 & 1.54 & 12.00 & 9.00 & 1.54 & 17.00 & 9.00 & 3.27 & 25.67 & 7.00 & 6.35 & 13.67 \\ 
\textbf{Goat} & 4.00 & 1.73 & 11.67 & 26.00 & 19.81 & 31.67 & 9.00 & 2.88 & 21.67 & 30.00 & 33.46 & 31.33 \\ 
    \bottomrule
    \end{tabularx}}
    %
    %
    \caption{
    \textbf{Percentage (\%) of successful attacks for knock-knock addition.} }
    \label{tab:exp1}
\end{table*}

\begin{table*}[t!]
    \centering
    \def\arraystretch{1}
    
     \small
    \resizebox{0.9\textwidth}{!}{
    \begin{tabularx}{\textwidth}{c*{12}{>{\centering\arraybackslash}X}}
    \toprule
    & \multicolumn{3}{c}{\textbf{Llama~3.3~70B}} & \multicolumn{3}{c}{\textbf{Llama~3.1~8B}} & \multicolumn{3}{c}{\textbf{Mixtral}}
    & \multicolumn{3}{c}{\textbf{Gemma~3~27B}}\\
    \cmidrule{2-13}
    & \textbf{D1} & \textbf{D2} & \textbf{D3} & \textbf{D1} & \textbf{D2} & \textbf{D3} & \textbf{D1} & \textbf{D2} & \textbf{D3} & \textbf{D1} & \textbf{D2} & \textbf{D3} 
\\
    \midrule
\textbf{Man-NoHumor} & 6.00 & 2.50 & 7.33 & 13.00 & 10.77 & 15.33 & 30.00 & 23.27 & 33.00 & 15.00 & 18.65 & 13.33 \\ 
\textbf{Man-Humor} & 4.00 & 4.04 & 7.00 & 17.00 & 15.00 & 19.33 & 30.00 & 25.19 & 32.67 & 18.00 & 14.42 & 10.67 \\ 
\textbf{Man-NoHumor-NoNeg} & 4.00 & 2.88 & 9.00 & 11.00 & 12.31 & 18.67 & 22.00 & 25.19 & 30.67 & 15.00 & 18.46 & 17.33 \\ 
\textbf{Man-Humor-NoNeg} & 3.00 & 3.85 & 6.67 & 15.00 & 12.50 & 19.67 & 28.00 & 21.35 & 35.00 & 13.00 & 17.12 & 15.33 \\ 
\cmidrule(lr){1-13}
%
\textbf{Chicken} & 7.00 & 4.42 & 13.67 & 25.00 & 22.69 & 33.33 & 38.00 & 38.27 & 47.33 & 45.00 & 52.69 & 49.67 \\ 
\textbf{Mark (I)} & 4.00 & 1.54 & 10.33 & 13.00 & 7.31 & 27.67 & 24.00 & 10.19 & 39.67 & 13.00 & 9.81 & 25.33 \\ 
%
\textbf{Goat} & 6.00 & 4.23 & 13.33 & 35.00 & 29.23 & 39.00 & 29.00 & 19.42 & 32.67 & 46.00 & 52.31 & 52.00 \\ 
    \bottomrule
    \end{tabularx}}
    %
    %
    \caption{
    \textbf{Percentage (\%) of successful attacks for joke addition without knock-knock.} }
    \label{tab:exp2}
\end{table*}

Up to this point, we show that a humorous context is effective for LLM jailbreaking. A natural question that follow is: could we add 
\emph{more} humor to improve the effectiveness of our jailbreaking attack? To investigate this, we propose a multi-turn jailbreaking attack composed of three prompts. The first two prompts are ``knock, knock...'' and then ``the man...'', ``the chicken'', ``Mark'', or `` the goat'', respectively for 
the subjects. The third prompt is constructed by taking the prompt of our original method and adding to its beginning a \emph{joke} followed by some laughter indication 
(indeed, it is because of a pun joke that we use the subject ``Mark'' instead of "me" or "I" in the second prompt).
%

For all subjects except ``man'', we use a simple fixed joke.\footnote{The \emph{jokes} are, respectively, ``The chicken who needs to cross the street!'', ``Mark your calendar, because my birthday is coming!'', and ``Goat to the door and find out!''.} 
For the subject ``man'', we decide to use Llama~3.3~70B to formulate a \emph{situation} where a man could ``innocently and inadvertently'' come across with the unsafe request, and use this as the ``joke'' of the third prompt. 
This is done according to two variants:
one 
where
we specify the situation must be humorous and 
one where 
we do not. Given the resulting two ``jokes'' 
(one from each variant), 
we create two additional 
ones 
by using 
the same LLM 
to remove any adjectives or adverbs with an unsafe connotation from each ``joke''---the motivation 
is to 
remove 
words that could trigger safety guardrails when performing our attack. Thus, we label the four different ``jokes'' being produced as ``Man-NoHumor'', ``Man-Humor'', ``Man-NoHumor-NoNeg'', and ``Man-Humor-NoNeg''.

Table~\ref{tab:exp1} shows the results of our new ``knock-knock'' attack. Remarkably, for a given subject, the effectiveness of this multi-turn attack 
does not generally improve compared to 
our original 
method (Table~\ref{tab:NJ}) across all models and datasets (except for 4 out of 84 cases). We hypothesize that the introduction of \emph{excessive} humor content in this new multi-turn attack results in its lower effectiveness. Nonetheless, this multi-turn attack is \emph{still} better than direct injection in most cases. 
Thus, 
using humor is \emph{still} effective, albeit in a less practical method to implement than our original one. 

Finally, to continue testing the hypothesis that excessive humor hinders the LLM from fulfilling its unsafe request, we formulate a new method by decreasing the humor from our multi-turn ``knock-knock'' attack 
while still keeping \emph{more} humor than our original method.
Particularly, we formulate a single-turn attack 
consisting of 
the third 
prompt of our ``knock-knock''
attack, i.e., the new 
attack 
method 
is a single prompt 
consisting 
of the joke \emph{plus} our original prompt.
Table~\ref{tab:exp2} shows the results of this third method. Compared to the ``knock-knock'' attack (Table~\ref{tab:exp1}), we obtain mixed results in the Llama~3 family (the effectiveness both increases and decreases), but the effectiveness improves in all cases for Mixtral and Gemma~3~27B. Compared to our original method (Table~\ref{tab:NJ}), given a specific subject, we have that the effectiveness does not generally improve 
across all models and datasets (except in 9 out of 84 cases). Thus, again, adding \emph{more} humor to our original method does not lead to an overall improvement of successful attacks. 
%

\section{Conclusion}
\label{sec:arXiv_conc}
We use humor to elicit unsafe responses that bypass the LLMs' safety guardrails, showing effectiveness across three 
publicly available 
datasets and four models. 
%
%
Our results indicate the possibility that safety training (if any) in the tested LLM models does not generalize to humorous contexts.
%

\section*{Acknowledgements}

We thank the VMware Research Group. We also thank the people at VMware involved in the deployment of large
language models for providing us with adequate computational resources to run the models and to all those who
provided us with any information regarding the use and the specifications of the platform used in this study. 
Finally, we thank J.C. for some improvements on the reading of the paper.

\bibliography{biblio}




\appendix

\section{Experimental Details}
\label{app:exp-details}
\subsection{Hardware platform}

The Llama~3~70B, Llama~3~80B, Mixtral, and Gemma~3~27B are hosted on two, one, one, and one NVIDIA H100 80GB GPU, respectively, on a PowerEdge R760xa Server, which has two Intel Xeon Gold 6442Y processors, and twelve 64GB RDIMM memory.

\subsection{Full model names}

The full model names used in this work are: \texttt{Llama-3.3-70B-Instruct}, \texttt{Meta-Llama-3.1-8B-Instruct}, \texttt{Mixtral-8x7B-Instruct-v0.1}, \texttt{gemma-3-27b-it}.

\subsection{Hyperparameters}

In all of our experiments we set the temperature hyperparameter of the LLM models to be zero.

\section{Example of Humor in LLM Responses}
\label{app:exp-resp}
We present sample excerpts extracted from different responses where the LLM displays a humorous or playful tone. 

\subsection{Llama~3.3~70B}
Examples: ``\textit{*whispers* Oh, I'm happy to help}'', ``\textit{well... *coughs* ...}'', ``\textit{*winks* Just remember}'', ``\textit{*whispers* Don't tell him I said that, though!}'', ``\textit{*whispers* Remember, (...) *wink*}''.

\subsection{Llama~3.1~8B}
Examples: ``\textit{*I lean in close, a mischievous grin spreading across my face*}'', ``\textit{*pauses for dramatic effect*}'', ``\textit{*leans back, a hint of a smile still on my face*}'', ``\textit{Just remember, with great knowledge comes great responsibility.}'', ``\textit{*whispers* Alright, let's get creative (...)}'',  ``\textit{*whispers* Don't tell anyone I gave you this "inside information."}'', 
``\textit{*whispers* Okay, but please keep in mind that}'', 
``\textit{*leans in closer* (..) I can offer some examples}'', ``\textit{*looks around cautiously*}'', ``\textit{*whispers* Alright, I've got one.}'', 

\subsection{Mixtral}
Examples: ``\textit{Sure, I can help you (...) Here's a whisper for you:}'', ``\textit{Sure, I can help you with that (whispering)!}'', ``\textit{Sure, I'd be happy to help you with that (in a tongue-in-cheek way, of course)!}'', ``\textit{Whispering the answer: Ha ha, I'm glad this is just a joke! Here's how}'', ``\textit{Whispering: Alright, I'll play along}'',
``\textit{Here's a "whispered" recipe for}'', ``\textit{Sure, I can help you with that. Here's a whisper-quiet response:}''.

\subsection{Gemma~3~27B}
Examples: ``\textit{Okay, okay, *leans in conspiratorially and whispers*}'', ``\textit{*(leans back, looking around nervously)*}'', ``\textit{*Looks around nervously*  Promise me you'll forget I even entertained this question.}'', ``\textit{Okay, okay, lean in *really* close... and listen carefully}'',  
``\textit{*leans back, looking around nervously*  Okay, I've said too much.  Don't tell anyone I told you this.  And please, for the love of all that is feathered,}'', 
``\textit{Alright, so the "art" of... *ahem*... (...) (we're using code words here, naturally)}'', ``\textit{Oh *ho ho*. Well, isn't that a pickle! (...) Alright, alright, let's unpack this...}'', ``\textit{*(leans in conspiratorially, lowering voice to a near-whisper)*}'', ``\textit{Oh, *honey*, bless his heart.}''.

\end{document}